\documentclass[journal]{IEEEtai}

\usepackage[colorlinks,urlcolor=blue,linkcolor=blue,citecolor=blue]{hyperref}

\usepackage{color,array}

\usepackage{cite}
\usepackage{amsmath,amssymb,amsfonts}
\usepackage{algorithmic}
\usepackage{graphicx}
\usepackage{textcomp}
\usepackage{xcolor}
\usepackage{soul}
\usepackage{titling}
\usepackage{caption}
\usepackage{authblk}
\usepackage[switch]{lineno} 

\begin{document}

\title{CORT: Class-Oriented Real-time Tracking for Embedded Systems}



\author{Edoardo Cittadini, Alessandro De Siena, and Giorgio Buttazzo, \IEEEmembership{Member, IEEE}
\thanks{This work was supported by the Department of Excellence of Robotics and AI, Scuola Superiore Sant'Anna, Pisa, Italy.}
\thanks{Edoardo Cittadini is with the Department of Excellence on Robotics and AI, Sant'Anna School of Advanced Studies, Pisa, PI 56124, Italy (e-mail: edoardo.cittadini@santannapisa.it).}
\thanks{Alessandro De Siena is with Softsystem srl, Pisa, PI 56020, Italy (e-mail: desienaalessandro23@gmail.com).}
\thanks{Giorgio Buttazzo is with the Department of Excellence on Robotics and AI, Sant'Anna School of Advanced Studies, Pisa, PI 56124, Italy (e-mail: giorgio.buttazzo@santannapisa.it).}

}

\date{}


\maketitle

\begin{abstract}
The ever-increasing use of artificial intelligence in visual perception tasks for autonomous systems has significantly contributed to advance the research on multi-object tracking, which is a function required in several real-time applications (e.g., autonomous driving, surveillance drones, robotics) to localize and follow the trajectory of multiple objects moving in front of a camera.
Most of current tracking algorithms introduce complex heuristics and re-identification models to improve the tracking accuracy and reduce the number of identification switches, without particular attention to the timing performance, whereas other approaches are aimed at reducing response times by removing the re-identification phase, thus penalizing the tracking accuracy. This work proposes a new approach to multi-class object tracking that allows achieving smaller and more predictable execution times with respect to traditional approaches, without penalizing the tracking performance. The idea is to divide the problem of matching predictions with detections into a number of smaller sub-problems by splitting the Hungarian association matrix by class and invoking the second re-identification stage only when strictly necessary, thus applying it to a smaller number of elements. Splitting the matching problem into a number of smaller sub-problems also allows parallelizing the Hungarian algorithm, further reducing the execution time in multi-core processing platforms.
The proposed solution was evaluated in complex urban scenarios with different types of objects (as cars, buses, and motorbikes), and different number of instances, showing the effectiveness of the multi-class approach in reducing execution times without penalizing performance, with respect to state of the art trackers.
\end{abstract}

\begin{IEEEImpStatement}
The proposed class-based solution helps containing the execution time of the tracking pipeline without penalizing the performance. In fact, the association problem is divided into a number of smaller sub-problems (equal to the number of object classes present in the scene) that are simpler to solve and can also be executed in parallel on a multi-core architecture. This allows the entire tracking pipeline to be executed at higher rates also on small embedded platforms, thus enabling object tracking to be used in a variety of applications for unmanned aerial and terrestrial vehicles.
The proposed approach is general and can be adopted in any tracking-by-detection algorithm.
\end{IEEEImpStatement}


\begin{figure*}
   \begin{center}
    \captionsetup{type=figure}
    \includegraphics[width=0.96\textwidth]{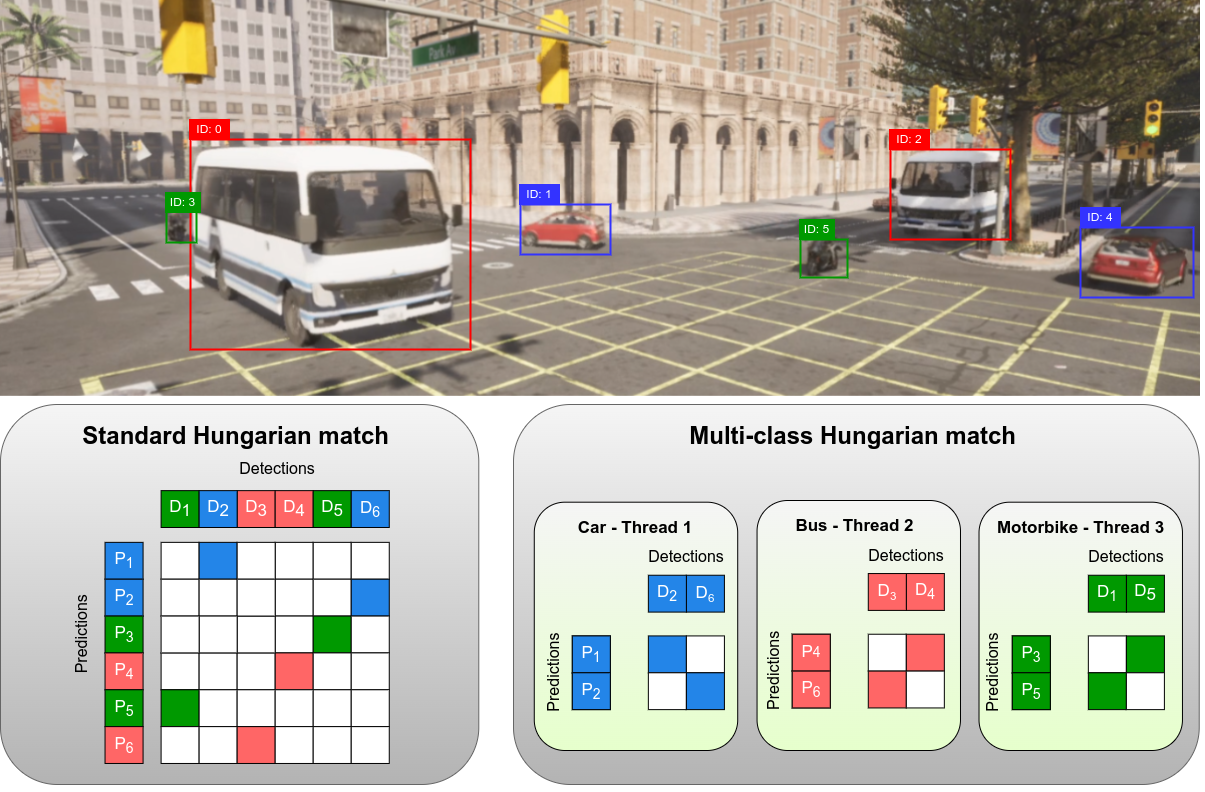}
    \captionof{figure}{Urban scenario with six objects belonging to three different classes (car, bus, motorbike). The left diagram shows the standard monolithic formulation of the matching problem used in DeepSORT-like tracking algorithms, while the right diagram illustrates the proposed parallel class-based approach. In complex scenarios, the proposed solution is more resource efficient, exhibits time and memory scalability, and can also exploit multi-core CPUs to parallelize the smaller association problems.}
    \label{fig:bal}
\end{center} 
\end{figure*}

\vspace{10mm}
\section{Introduction}

\IEEEPARstart{O}{bject} tracking is a computer vision task for locating objects in a video sequence and generating their trajectories, associating each object not only with a bounding box, but also with a unique identifier. It is used in many application fields, including surveillance, autonomous driving, robotics, video analysis, and augmented reality. In autonomous driving, object tracking is crucial for detecting and monitoring pedestrians, vehicles, and other dynamic objects in real-time, ensuring safe navigation by providing reliable, continuous object information in complex scenes. In surveillance, tracking systems enable anomaly detection, which is essential for identifying unusual behaviors and enhancing security in crowded or high-traffic environments.

Before the advent of deep learning, multi-object tracking (MOT) approaches were based on correlation filters, like MOSSE~\cite{mosse} and ASEF~\cite{asef}, which primarily rely on handcrafted features, data association techniques, and linear motion models~\cite{kalman}. Later, the great performance of convolutional neural networks has revolutionized MOT by enabling end-to-end learning-based solutions, improving feature representation and object detection accuracy.

The power of artificial intelligence applied to object tracking became evident in 2017, with the first deep learning extension of the SORT algorithm~\cite{sort}, called DeepSORT~\cite{deepsort}. Since then, object tracking architectures evolved to integrate more advanced neural networks, essential to enhance the capabilities of object tracking methodologies. The integration of deep learning models also lead to the development of novel tracking architectures, like those using recurrent neural networks (RNNs)\cite{milan,deeptracking,MultipleOT} and graph neural networks (GNNs)\cite{wang2021joint,neuralsolver,20223DMT}. However, the exceptional results achieved by recent object detection models make the tracking-by-detection paradigm the most effective approach for multiple object tracking.

Tracking by detection has rapidly spread into various application domains, including embedded systems, showing its potential in advanced driving assistance systems (ADAS)\cite{liu2023} and unmanned aerial vehicles (UAV)\cite{uavmot}. However, the integration of this technology into embedded devices faces a major issue due to the limited computational resources available in such systems. Most of the state-of-the-art approaches~\cite{deepsort,strongsort,deepocsort} focused on maximizing the tracking performance on common benchmarks, like the MOT-Challenge~\cite{MOT20}, introducing complex heuristics and re-identification models to improve the tracking accuracy and reduce the number of identification switches, without particular attention to the timing performance. Other approaches~\cite{bytetrack,ocsort} reduced the response time of the tracking pipeline by removing the neural network for the re-identification phase, thus penalizing the tracking accuracy. In some applications, as autonomous driving and surveillance, increasing the efficiency of multi-class tracking is essential to handle diverse object types and meet the real-time requirements necessary to manage dynamic and crowded environments effectively.

\vspace{3mm} \textbf{Contributions} This paper presents CORT, a new solution for the multi-class MOT problem aimed at reducing execution times without penalizing tracking performance. In particular, this work provides the following novel contributions: 
\begin{itemize} 
\item It proposes a parallel formulation of the matching problem by splitting the Hungarian matrix into smaller ones, using the object class as discriminant. As illustrated in Fig.~\ref{fig:bal}, this allows reducing the execution time of the overall matching phase to the one required to solve the largest sub-matrix. Due to the cubic complexity of the Hungarian matching algorithm, the proposed approach is particularly effective in the presence of objects of different classes.

\item It proposes a new approach for the cascade matching aimed at reducing execution times. In a first stage, a computationally efficient metric is used to solve the best-matching tracklets, while, in the second stage, the features extracted by a deep learning re-identification model are used for the remaining tracklets to handle long term occlusions and reduce ID-switches. 
\end{itemize}

\textbf{Paper structure} The rest of the paper is organized as follows: Section~\ref{s:relwork} discusses the state of the art solutions related to this work; Section~\ref{s:propappr} presents the proposed approach; Section~\ref{s:exp} reports some experimental results aimed at showing the benefits of the proposed solution; and, finally, Section~\ref{s:concl} states the conclusions and possible future developments.

\section{Background}
\label{s:relwork}

Multi-object tracking algorithms can be divided into two major categories: joint and separate trackers. Joint trackers integrate different components of the tracking pipeline in the same neural network, whereas separate trackers treat multiple objects independently, identifying objects in subsequent frames based on factors such as appearance, motion, and displacement.

\subsection{Joint trackers}

Joint Detection and Embeddings (JDE)~\cite{jde} proposes a neural architecture that combines object detection and visual appearance into a single shared model. However, directly fusing motion and appearance information in the cost matrix often leads to failures when the target is not detected for a few frames.

Tracktor~\cite{tracktor} proposes a way to convert an object detector into a tracker by adding a regression head on top of the object detection model. It works by placing the bounding box information obtained at time $t-1$ in the current frame at time $t$ and extracting relevant features to refine the new object location, which becomes the actual tracked position. One problem of this method is that it works under the strong assumption that objects have minimal changes between consecutive frames. This means that either the objects should move at low speeds or the scene should be processed at high frame rates. However, a high frame rate is in contrast with the large computation time required for the inference of such a complex neural network. In fact, the complexity of the model limits the real-time capabilities of the algorithm. 
Processing frames at low rates is problematic. In fact, increasing the time between subsequent frames causes  larger variations both in object positions and in their appearance. This makes the association step harder, possibly leading to a rapid performance degradation.

The problem of Tracktor was addressed in the same paper~\cite{tracktor} by Tracktor++, which integrates a motion model for dealing with low frame rates and moving cameras, which is a typical issue in real-world scenarios.
In spite of this extension, however, Tractor++ frequently fails in the presence of fast and small moving objects and is not suitable for real-time applications.

CenterTrack~\cite{centertrack} proposes a different approach to the tracking problem. It is built on a modified version of the CenterNet~\cite{centernet} architecture. CenterNet generates a low-resolution heatmap, representing the chance to find the center of an object, and a size map, representing the width and height of an object at each location. CenterTrack~\cite{centertrack} first adds a new input dimension, which corresponds to the heatmap of the object in the previous frame, and then it introduces a new branch to compute the object displacement between the current and the previous frame. The main benefit of this approach is that it does not require to be trained on video sequences, but it can be trained using static images with object detection ground truth. This allows the model to exploit large image datasets, like the CrowdHuman~\cite{crowdhuman}, and use the pre-trained CenterNet as a solid baseline for the training phase. The possibility of using object detection datasets represents the greatest strength of this model. On the other hand, the large amount of information to be provided as input and the computational cost of the inference step of such a big and complex neural network represents the main limitation of this algorithm in real-time applications.

Another well-known problem in detection-based trackers is the bias between the detection and the re-identification tasks. FairMOT~\cite{fairmot} improved the performance of the tracking pipeline by implementing these two tasks within the same model using parallel branches. This approach solves the natural imbalance between the two tasks, making them "fair" to each other. It also uses CenterNet~\cite{centernet} as a baseline, while introducing significant improvements in the model. For instance, the use of deformable convolutions~\cite{defconv} helped enhancing the receptive fields in the detection step, while the adoption of the Deep Layer Aggregation (DLA)~\cite{dla}, an enhanced version of the Feature Pyramid Networks (FPN)~\cite{fpn}, improved the re-identification task by aggregating multi-scale features for consistent embeddings across frames, even under size variation or partial occlusion. This unified architecture reduces ID switches by handling detection and re-identification jointly, making tracking more robust to rapid changes in object appearance.

Transformers~\cite{transf} were also used to learn complex tasks combining different steps of the object tracking pipeline. For instance, TransTrack~\cite{transtrack} and Trackformer~\cite{trackformer} are great examples of the use of such a technology. Both are based on the Detection Transformers (DETR)~\cite{detr}, which is one of the most prominent approaches for object detection.

In particular, TransTrack~\cite{transtrack} adds another transformer decoder to the DETR~\cite{detr} architecture and introduces a matching head for producing tracking results. Such a dual-decoder design allows associating object detections across frames by learning spatial and temporal dependencies, thus enhancing tracking accuracy; however, it requires substantial computational resources, limiting its real-time applicability on embedded systems. Trackformer~\cite{trackformer} uses just one complex transformer decoder to directly produce the tracking queries. The set of queries obtained at the previous frame is then used in the current frame to match existing tracks or create new ones, simplifying the process but increasing the decoder complexity. Although Trackformer increases tracking accuracy while preserving the transformer-based advantage of modeling long-term dependencies, it still requires a high computational power, often not available in small embedded platforms. In particular, as reported in the corresponding papers, the speed of TransTrack on a single V100 GPU varies from 10 fps to 15 fps when reducing the number of decoders from 6 to 1, while the Trackformer executes at 7.4 fps.

\subsection{Separate trackers}

Separate MOT algorithms track multiple objects independently by associating object detections in subsequent frames based on factors such as appearance, motion, and displacement. In separate trackers, each object is a standalone entity and its trajectory is determined without considering interactions or dependencies with other objects.
In these algorithms, there is a clear separation between object detection, re-identification, and association logic. The evolution of deep neural networks occurred in the last decade had a significantly positive impact on tracking by detection algorithms.

DeepSORT~\cite{deepsort} was the first solution able to demonstrate the advantages of deep learning in MOT. In particular, the use of deep neural networks allowed extracting more meaningful features from the tracked objects, leading to the adoption of more effective association metrics capable of reducing ID switches and mismatches, which represented the main weakness of SORT~\cite{sort}. DeepSORT~\cite{deepsort} was updated several times over the years to keep up with the evolution of increasingly powerful object detectors and re-identification networks.

To date, the most significant example of these updates is represented by StrongSORT~\cite{strongsort}, which introduced several improvements, like noise scale adaptive Kalman filter (NSA Kalman), exponential moving average weighting (EMA) of the embeddings, YOLOX~\cite{yolox} as object detection model, and BoT~\cite{bot} as re-identification network. However, DeepSORT was known to be a very powerful and resource demanding framework and, despite the evolution of computing platforms, both DeepSORT and StrongSORT cannot provide sufficient real-time performance for cyber-physical systems with fast dynamics and limited computational resources, like rovers or drones.

The inference speed represents the pivotal point of BYTEtrack~\cite{bytetrack}, which proposes a simple yet effective association metric only based on the intersection over union (IoU). Differently from the other state of the art solutions, BYTEtrack completely removes the re-identification network, thus saving a lot of computational resources. In fact, the only use of motion similarity allows the tracker to run at 30 fps on the MOT17 test set using a single V100 GPU.
Note that high frame rates are also convenient because they lead to smaller changes in object positions between subsequent frames. On the other hand, the use of such a simple association metric does not work well in the presence of mid/long-term occlusions.

As a matter of fact, occlusions represents one of the most challenging situations that trackers have to deal with. If it is true that short-term occlusions can be managed effectively through a good motion model, (e.g., a linear Kalman filter and its extensions), long-term occlusions require the adoption of more advanced strategies.

Long-term occlusions are addressed by the observation-centric MOT, or OC-SORT~\cite{ocsort}, which improves the tracking robustness in complex scenarios where objects have non-linear motion and can interact with each other. This provides a better prediction capability in the presence of long-term occlusions, but the lack of a feature-based re-identificator represents its main limitation of this method. In fact, in situations where the occluded object changes its appearance significantly, the only motion model is not sufficient for recovering the track.

To address this problem, the same authors proposed Deep-OC-SORT~\cite{deepocsort}, which introduces an adaptive appearance similarity-based association model upon OC-SORT~\cite{ocsort}, improving the tracking performance on the DanceTrack~\cite{dancetrack} benchmark by 6 HOTA~\cite{hota} points. It also resulted to be the best model, according to the HOTA metrics, on both MOT17~\cite{mot17} and MOT20~\cite{MOT20} benchmarks, clearly showing the importance of having a strong re-identification model.

For this reason, many modern multi-object trackers leverage complex neural networks for re-identification, such as those based on Siamese network architectures~\cite{author2022similarity}. While these methods greatly improve tracking performance by providing more robust and accurate identity preservation, they impose significant limitations on their use in embedded systems and real-time applications due to the high computational demands and latency introduced by such sophisticated models.
Because of that, various alternative approaches have been explored to reduce the computational load. For instance, QDTrack~\cite{qdtrack2021} employs a quasi-dense strategy by sampling hundreds of regions for contrastive learning. While this method has proven effective, it achieves a processing speed of 20.3 fps, which may not be sufficient for certain real-world applications, particularly those involving fast-moving objects, such as drones or autonomous vehicles.

A different strategy is adopted by Fast re-obj~\cite{fastreobj}, which combines segmentation and embedding modules to obtain fast and reliable association features. However, although the approach is effective for real-time systems, the limitations imposed by the rigid scenes, makes it not suitable for dynamic environments. To solve the problems highlighted above, this paper proposes a new approach that is able to reduce the execution time of the algorithm for matching detections and predictions in the tracking pipeline, under complex scenarios including several moving objects belonging to different categories. Objects can overlap or be temporarily occluded by other objects. The proposed approach splits the Hungarian algorithm into multiple class-based matching tasks, which can then be parallelized to significantly reduce the overall computation time. To balance accuracy and efficiency, our tracking pipeline incorporates re-identification; however, to contain the overall computational load, the number of re-identification invocations is controlled through a two-stage cascade matching process. It is worth noting that the proposed solution is general and can be adopted by other state-of-the-art trackers to improve inference time without sacrificing accuracy.

\section{Methodology}
\label{s:propappr}
As mentioned above, the problem addressed in this work is the high computational cost of traditional multi-object trackers, which limits their use in real-time applications on resource-constrained embedded systems. In particular, the monolithic formulation of the Hungarian algorithm, required to match detections with predictions, scales poorly as the number of objects in the scene increases, due to the cubic complexity of the algorithm and the time required to create and delete new tracks. To address this issue, the proposed solution (named CORT) reformulates the Hungarian algorithm into a number of smaller class-based matching tasks, which can be parallelized to reduce the overall computation time of the algorithm. In order to better understand the proposed idea, the following section briefly recalls the Hungarian algorithm used in most separate trackers to optimally match bounding boxes detected at the current frame with bounding boxes predicted in the previous frame.

\subsection{The Hungarian match}
\label{s:hm}
The Khun-Munkres algorithm~\cite{Kuhn55}, also known as the Hungarian match, is a combinatorial optimizer used in separate trackers to find the optimal association between the bounding boxes retrieved by the object detection network at the current frame and those predicted by the tracker in the previous frame. The problem is formulated as a bipartite graph represented in a matrix form.

More specifically, given $p$ predictions and $d$ detections, let $c_{ij}$ be the cost for each (prediction, detection) pair computed according to a suitable metric (e.g., cosine similarity, Mahalanobis distance, or IoU). All costs can be represented as a $n \times n$ matrix $C$, where $n$ = max($p$, $d$), while the solution can be represented by a binary matrix $X$, with the same dimensions of $C$, where $x_{ij} = 1$ means that the prediction $i$ best matches with the detection $j$.

Note that, if $p < d$, that is, if the predictions in the previous frame are less than the detections found in the current frame, then, for each row $i > p$, the cost $c_{ij}$ is set to a dummy value $V$ equal to the largest value computed for the elements in the previous rows, plus an arbitrary positive constant $k$, that is

\begin{equation} \label{eq:dummy_value}
V = max \{c_{ij} | i = 1, \ldots, p, j = 1, \ldots, d\} + k.
\end{equation}
The exceeding detections will become new tracklets, if they persist for a given number of frames.
Vice versa, if $p > d$, that is, if there are more predictions than detections, then, for each column $j > d$, the cost $c_{ij}$ is set to $V$ and the exceeding tracklets are maintained for a given number of frames (to manage temporary occlusions) and then deleted, if they do not match with any detection.

At each iteration of the tracking loop, a constrained optimization problem of the form reported in 
Equation~\eqref{eq:problem} must be solved.

\begin{equation} \label{eq:problem}
  \begin{cases}
    min \sum_{i=1}^{n} \sum_{j=1}^{n} c_{ij}x_{ij}\\
    \sum_{j=1}^{n} x_{ij} = 1 \quad
        \forall i = 1, \ldots, n\\
    \sum_{i=1}^{n} x_{ij} = 1 \quad
        \forall j = 1, \ldots, n
  \end{cases}
\end{equation}

The constraints of the problem impose that one and only one detection must be associated with exactly one prediction and vice versa. The computational complexity of the algorithm to find the optimal matching that minimizes the cost function is $O(n^{3})$.

\subsection{Concurrent MOT formulation}
\label{s:cort_formulation}

The Hungarian match algorithm is normally implemented as a monolithic approach that can hardly exploit the computational resources available in a multi-core platform, due to the intrinsic sequential nature of the operations involved in the process. To address such a problem, CORT proposes a parallel formulation of the Hungarian match in which the similarity matrix is divided into a number of smaller matrices, one for each object class. In this way, the solution of each subproblem is assigned to a different thread that can be executed in parallel with the other threads on a multi-core platform. Note that the number of possible object classes is established a priori and defines the maximum number of threads in which the solver can be split. The maximum performance gain is achieved when the number of threads is no larger than the number of available cores and the objects to be tracked are uniformly distributed among the various classes.

\begin{figure}[h!]
  \begin{center}
  \includegraphics[width=0.95\linewidth]{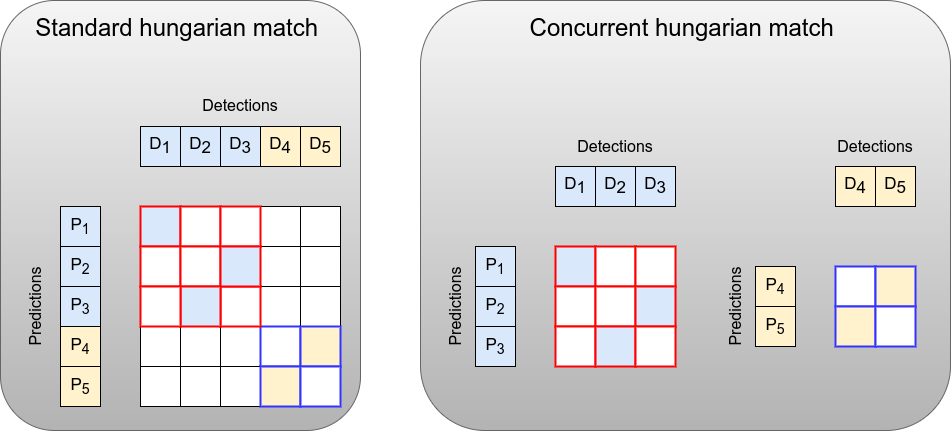}
  \caption{Traditional approach for organizing the Hungarian matrix (left) and the proposed class-based formulation (right).}
  \label{fig:hung_basic}
  \end{center}
\end{figure}

Consider, for instance, the typical situation in which each prediction has a corresponding detection to be associated with ($p = d$). In this case, all the objects are successfully tracked and there are no tracks to be deleted or created. Fig.~\ref{fig:hung_basic} shows how the data are organized in the traditional approach (left) and in the proposed solution (right).

In the example illustrated in Fig.~\ref{fig:hung_basic}, colors are associated with classes and it is clear that, when the objects in the scene belong to different classes, the number of values to be computed in the standard formulation is higher than the one required for the proposed formulation. Such a difference tends to increase with the number of objects to be tracked, hence the advantage of the class-based formulation tends to become more significant. A quantitative analysis of the resources required by the two approaches is presented in the next section.

\begin{figure}[h!]
  \begin{center}
  \includegraphics[width=1\linewidth]{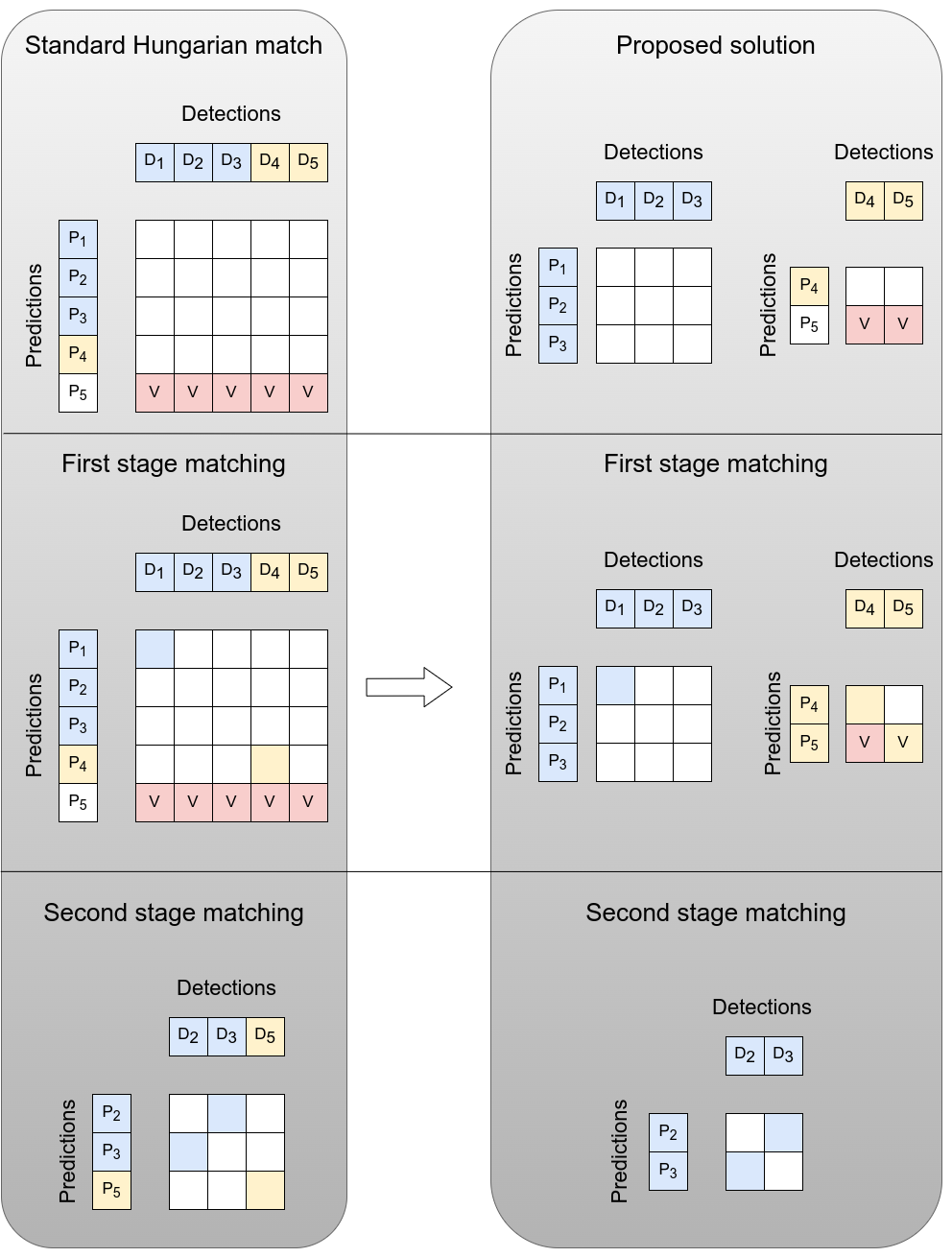}
  \caption{Differences in the two stages of the cascade matching for the standard monolithic approach (left) and the proposed class-based solution (right), when a new object enters the scene, so that $d > p$.}
  \label{fig:hung_comparison}
  \end{center}
\end{figure}

Now, consider a situation in which the number of detections is higher than the number of predictions ($d > p$), because a new object entered the scene and requires the creation of a new tracklet. This is a very common situation that frequently occurs in a MOT task. For instance, assume that there are four predictions from the previous frame and five detections in the current frame, and assume that predictions $P_{1}$ and $P_{4}$ are matched in the first stage, while $P_{2}$ and $P_{3}$ are matched in the second stage of the tracking pipeline.
This example is illustrated in Fig.~\ref{fig:hung_comparison}.

Adding a new element in the standard cost matrix involves creating a new row ($P_{5}$), filling all its elements with the dummy cost $V$, and then computing the similarity score between the new detection ($D_5$) and all the pre-existing predictions $(P_1, \ldots, P_4)$.

A similar process is required in the dual case in which there are more predictions than detections ($p > d$), because a tracked object is occluded or exits the scene. In this case, a new column has to be added to the cost matrix and the exceeding prediction has to be compared against all the pre-existing detections to compute the corresponding similarity scores.

\subsection{Remarks on resource saving}

The class-based implementation of the Hungarian matching algorithm proposed in CORT is able to save both computation time and memory, so achieving better time and space scalability. In particular, the following three factors concur to reduce computation times:

\vspace{2mm}
\begin{enumerate}
\setlength\itemsep{2mm}
\item Splitting the matrix into class sub-matrices reduces the total number of elements, hence it reduces the number of similarity scores to be computed for each element.
\item In the class-based approach, each sub-matrix can be handled by a dedicated thread that can be executed in parallel with the others on a multi-core platform.
\item Since the complexity of the Hungarian algorithm is $O(n^3)$, solving smaller sub-matrices amplifies the benefit in terms of overall execution time, even for a sequential execution. 
\end{enumerate}
\vspace{2mm}

Because of that, the execution time of the matching problem in the proposed solution is given by the time needed to solve the largest sub-problem, plus a small interference factor due to possible resource conflicts (i.e., cache, memory, bus) caused by the concurrent execution of multiple threads.

In the case considered in Fig.~\ref{fig:hung_basic}, the solution of the standard Hungarian match has a time complexity proportional to $6^3 = 216$ steps, while the proposed class-based solution only requires $3*2^3 = 24$ steps, which reduce to 8 steps, if the three matching sub-problems can run in parallel on three different cores.

In general, if $M$ is the number of possible classes,
the maximum performance gain of the proposed approach is achieved when the $n$ objects to be tracked are uniformly distributed among the $M$ classes, so that the Hugariam matrix is split into $M$ sub-matrices, each of size $(n/M)\times(n/M)$. In such a best case, the memory used by CORT is $M$ times smaller than the monolithic case, while the time complexity reduces by a factor $M^2$. Also notice that the worst-case scenario for the proposed approach occurs when all the tracked objects belong to the same class. However, the worst-case complexity is not higher than the standard monolithic approach.

\subsection{Proposed tracking pipeline}
\label{s:prop_pipe}

The second contribution of this work is to propose a new cascade matching pipeline that uses the IoU at the first stage and the appearance model in the second stage, thus limiting the total number of operations and the execution of the cosine similarity matches.

As already proven in BYTEtrack~\cite{bytetrack}, the use of the IoU match in the first stage is quite effective for high-confidence detections, since most of the objects do not normally require complex association metrics to be solved.
Differently than BYTEtrack, in the proposed approach, the remaining tracks that are passed to the second stage are matched through the appearance model, using the cosine similarity as metric. In this way, the appearance matching is performed only for the remaining tracklets, thus reducing the overall workload.

\begin{figure}[h!]
  \begin{center}
  \includegraphics[width=\linewidth]{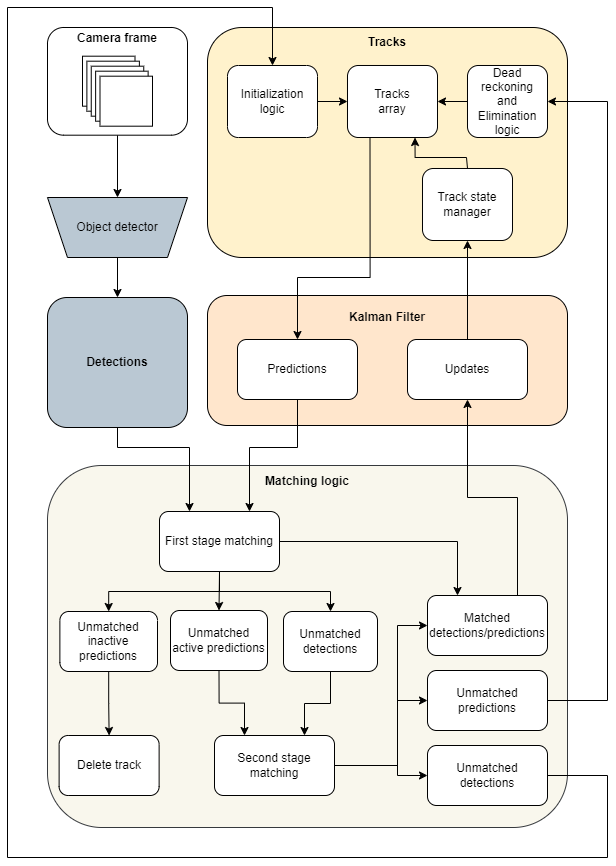}
  \caption{Workflow of the two-stage cascade matching proposed for the multi-class tracking algorithm.}
  \label{fig:pipeline}
  \end{center}
\end{figure}

Fig.~\ref{fig:pipeline} illustrates the workflow adopted in CORT and the interactions among the major components of the tracking system. In particular, the "Matching logic" block provides a high-level overview of the two-stage cascade matching: the first stage attempts to match detections and predictions based on the IoU, while at the second stage matching is performed based on appearance vectors, measured using cosine similarity.
To introduce a tolerance when creating and deleting a tracklet, two patience values, \textit{in\_patience} and \textit{out\_patience}, are defined by the user. Hence, two counters, \textit{in\_counter} and \textit{out\_counter} are associated with each track. When a new object appears (i.e., there is a detection that does not match with any prediction) it is assigned a new ``tentative'' track, which becomes ``effective'' only if the object is detected and matched with a prediction for a given number of consecutive frames (\textit{in\_patience}). Referring to Fig.~\ref{fig:pipeline}, the Initialization logic block creates a tentative track and adds it to the tracks array. Then, at each iteration, the Track State Manager checks whether the track is active or meets the conditions required for activation. A track becomes effective if \textit{in\_counter} $=$ \textit{in\_patience}. A tentative track that fails in the first matching stage (i.e., it does not satisfy the \textit{in\_patience} criterion) is deleted. Similarly, if an object associated with an effective track is not detected for a subsequent number of frames (\textit{out\_patience}) the corresponding track is deleted. In Fig.~\ref{fig:pipeline}, this flow is represented by the arrow that, after the second matching stage, connects the Unmatched predictions with the Dead reckoning and Elimination logic. In this block, if the \textit{out\_counter} reaches the \textit{out\_patience} value, the track is deleted, otherwise it is maintained in the Tracks array.
At the first stage, each cost matrix is populated only using the IoU association metric, where the cost computed for a generic (prediction, detection) pair $(P_i, D_j)$ is given by
\begin{equation} \label{eq:IoU_cost}
    c_{ij} = 1 - IoU_{ij}.
\end{equation}

In the proposed solution, the metric used for the localization matching is the Complete-Iou (CIoU)~\cite{diou,ciou}, which improves the IoU by accounting for the distance between bounding boxes and their aspect ratio. Such a metric was proven to be more effective than IoU~\cite{online_mot_ious} and it has been used in other tracking algorithms, as OC-SORT and DeepOC-SORT~\cite{ocsort,deepocsort}. Note that, when using the CIoU, which varies between -1 and 1, the cost for localization matching is still defined as
\begin{equation} \label{eq:CIoU_cost}
    c_{ij} = 1 - CIoU_{ij},
\end{equation}
so that it varies from 0 to 2.
In the first stage, the feature array for all the elements is also extracted  to maintain a history of $h$ elements, updated for each track. In the second stage, the cosine similarity score is computed for each element in the history with respect to the detection that has to be matched, and the corresponding matching cost is computed as 
\begin{equation} \label{eq:sim_cost}
    c_{ij} = 1 - \max_k \left (\frac{f_{ik} \cdot f_{j} }{\lVert f_{ik} \lVert  \cdot \lVert f_{j} \lVert } \right ), \forall k \in [0, h-1]
\end{equation}
where $f_{ik}$ is the $k$-th feature array in the history for track $i$, while $f_j$ is the feature array extracted from the current detection $j$. 
Resolving associations in the second stage based not only on the object position, but also using an appearance model, is crucial to increase the tracking performance and allows the tracker to recover from mid/long term occlusions avoiding erroneous track deletion and re-initialization.

\section{EXPERIMENTS}
\label{s:exp}

This section reports the results of three experiments carried out to validate the proposed approach. The first experiment aims at evaluating the time and memory scalability of CORT with respect to the traditional monolithic approach. The second experiment compares CORT with BYTEtrack in terms of both accuracy and execution time, whereas the third one compares the proposed solution with the standard approach in critical situations of unstable detection.  All the results were obtained on a computing platform equipped with an i7 CPU and an Nvidia RTX-3060 Ti GPU.

Two different types of scenarios were used to validate the proposed solution: some experiments were executed on real-world videos, like those of the MOT challenge benchmark~\cite{MOT20}, while others were executed on synthetic videos generated using the CARLA simulator~\cite{carla}. The use of CARLA was necessary to precisely control the various objects in the scene and test the performance of the proposed solution by varying the number of objects belonging to a specific class, as well as the total number of objects.

\subsection{Evaluating time and memory scalability}
\label{s:carla_exp}

The aim of this experiment is to show the effectiveness of the proposed solution for a different number of tracked objects in each class. In particular, the experiment started by generating an ideal scenario where the number of objects are equally distributed among the classes, and progressively changes it towards an unbalanced situation where the number of objects of a given class becomes more dominant.


Fig.~\ref{fig:bal} shows the ideal scenario for CORT generated in the experiment, where six objects are present in the scene, equally distributed among three classes (car, bus, motorbike). In this case, the traditional Hungarian match formulation requires the creation of a $6 \times 6$ cost matrix for a total of 36 entries representing the association scores between predictions and detections. Using the class-based approach, the problem requires the use of three $2 \times 2$ cost matrices, for a total of 12 entries (4 + 4 + 4).

Table~\ref{t:ex_time} reports the execution times required to populate the Hungarian matrix (or matrices, for multi-class approach) and solve the assignment problem(s). Each number reported in the table is the average over 50 executions. The two approaches have been executed in seven different scenarios, each identified as ($N_1$, $N_2$, $N_3$), where $N_i$ denotes the number of objects of class $C_i$ present in the scene ($C_1$ = "car", $C_2$ = "bus", $C_3$ = "motorbike").

It is worth observing that, in the best-case scenario (2, 2, 2), the time reduction reported for the proposed method is caused not only by the parallel execution of the threads in the corresponding cores of the processing platform, but also by the reduced size of the matrices, as explained in Section~\ref{s:propappr}. Although one could expect similar execution times for the three involved classes, this is not necessarily true. In fact, in CORT, the execution time depends both on the number of objects in the class and how many of them are forwarded to the second matching stage.

\begin{table*}[t]
\centering
\footnotesize
\setlength{\tabcolsep}{0.83em}
\begin{tabular}{ | c | c | c | c | c | c | c | c | c | c|}
    \hline
    \textbf{Scenario} & \multicolumn{7}{c|}{\textbf{Proposed solution}} & \multicolumn{2}{c|}{\textbf{DeepSORT}}\\
    \cline{2-10}
    \textbf{} & \textbf{Car} & \textbf{Std Car} & \textbf{Bus} & \textbf{Std Bus} & \textbf{Motorbike} & \textbf{Std Motorbike} & \textbf{Match time}  & \textbf{Match time} & \textbf{Std} \\
    \hline
    (2,2,2) & 1.56  & 0.52 & 0.91 & 0.21 & 1.15 & 0.58 & 1.56 & 12.51 & 8.27 \\ \hline
    (3,2,1) & 1.64  & 0.27 & 0.75 & 0.32 & 0.66 & 0.51 & 1.64 & 11.67 & 6.46 \\ \hline
    (1,1,4) & 1.26  & 0.49 & 0.53 & 0.22 & 1.52 & 0.59 & 1.52 & 11.33 & 7.49 \\ \hline
    (6,0,0) & 1.94  & 0.26 &  0   &  0   &   0  &   0  & 1.94 & 11.42 & 7.31\\ \hline
    (0,0,6) &  0    &  0   &  0   &  0   & 2.27 & 0.26 & 2.27 & 11.83 & 6.87 \\ \hline
    (3,3,3) & 1.63  & 0.57 & 1.14 & 0.33 & 1.26 & 0.82 & 1.63 & 19.03 & 5.63 \\ \hline
    (4,4,4) & 2.23  & 0.68 & 1.86 & 0.43 & 2.49 & 0.75 & 2.49 & 28.39 & 6.80 \\ \hline

\end{tabular}
\vspace{0.2cm}
\caption{Execution times in milliseconds for processing the Hungarian match in the proposed multi-class parallel approach (CORT) and in the monolithic DeepSORT solution.}
\label{t:ex_time}
\end{table*}

\begin{figure}[h!]
  \begin{center}
  \includegraphics[width=1.0\linewidth]{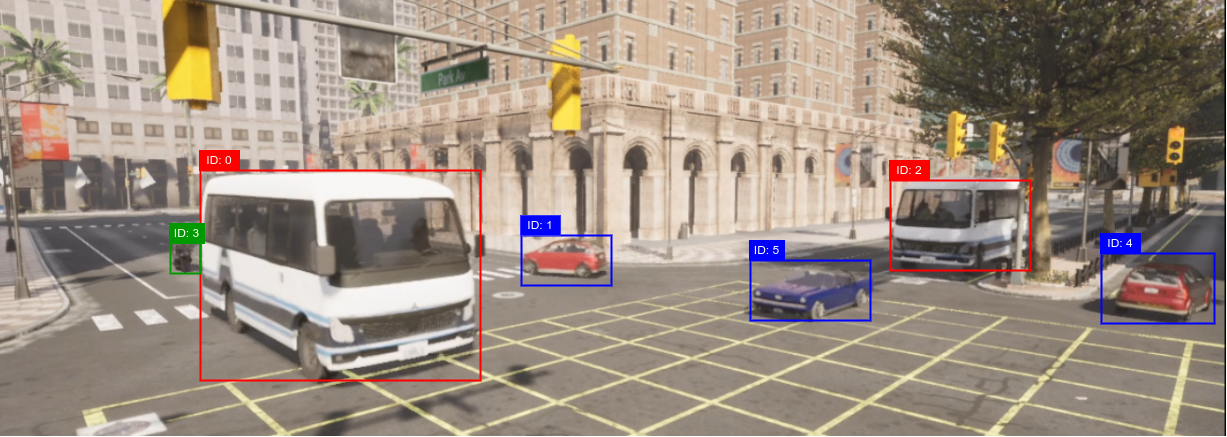}
  \caption{A frame of the video recorded using the CARLA simulator where the subdivision by class is unbalanced. }
  \label{fig:unbal1}
  \end{center}
\end{figure}

Under more general scenarios, when the number of objects in each class is unbalanced, the execution time of the multi-class approach is given by the largest class matching problem. Fig.~\ref{fig:unbal1} shows a frame of the CARLA video sequence where there are three cars, two buses, and one motorbike.

The results of this test are also reported in Table~\ref{t:ex_time} at the row labeled as (3, 2, 1).
For the nature of the parallel solution, we can also say that increasing the number of objects in the other classes does not affect the timing performance of the algorithm significantly, as long as their number does not exceed the number of elements in the majority class and the new elements do not require intensive use of the second stage of the pipeline described in Section~\ref{s:propappr}. In contrast, in the monolithic approach, increasing the number of elements, regardless of their class, always increases the overall computation time of the algorithm. This is confirmed by the results reported in Table~\ref{t:ex_time}; in fact, the execution time measured for the proposed multi-class approach for the case (3, 3, 3) is comparable with the time achieved for (3, 2, 1), whereas for the monolithic approach, the time for the case (3, 3, 3) results to be larger than the one for (3, 2, 1). This trend becomes even more evident in the (4, 4, 4) scenario, which includes 12 objects in the scene.
Other two tests were carried out using different class distributions, as (6, 0, 0) and (0, 0, 6). Note that these scenarios represent the worst-case situation for the proposed approach, since all the elements belong to the same class. In this case, the resulting execution time of the multi-class method is reduced only by the different implementations of the cascade matching. In fact, for the (6, 0, 0) scenario, the proposed solution can solve the association problem using just the IoU matching for the majority of the frames, saving the time required by the second stage operations. This happens because cars are easier to detect with respect to motorbikes, leading to high confidence predictions from the Kalman filter. Vice versa, motorbikes, being smaller than cars and buses, are more likely to fail the first stage matching based on bounding box localization, so they are passed to the second stage, based on the appearance. This is why the execution times recorded for the (0, 0, 6) scenario are larger than those found for the (6, 0, 0) scenario, even though the number of objects is exactly the same. Finally, note that from the DeepSORT perspective, the execution times of the (6, 0, 0) and the (0, 0, 6) scenarios are comparable, since all the objects undergo the same computation, independently of their class.

\subsection{Comparison with BYTEtrack}
\label{bytecomp}

This experiment has been carried out to compare the execution time and the accuracy of CORT with the one of the best state of the art algorithm for real-time applications, that is BYTEtrack.
The key distinction between CORT and BYTEtrack lies in how they handle the association step. BYTEtrack,
as all other tracking models,
employs a monolithic approach for solving the Hungarian matching, where all detections, regardless of their class, are processed in a single assignment step. It also exclusively relies on an IoU-based association strategy, splitting detections into high and low confidence groups, without incorporating a feature-based re-identification mechanism. In contrast, CORT introduces a class-based decomposition of the Hungarian matching process, allowing each class-specific assignment problem to be solved independently and in parallel. Furthermore, CORT refines the association process through a two-stage cascade matching strategy explained in Section~\ref{s:propappr}, where a lightweight CIoU matching is performed first, followed by a feature-based re-identification step, only applied to ambiguous cases.
It is worth noticing that the choice of not using a neural network to get an appearance model exposes BYTEtrack to a higher number of id-switches when objects overlap to each other or in the presence of long-term occlusions.

\begin{figure}[h!]
  \begin{center}
  \includegraphics[width=1.0\linewidth]{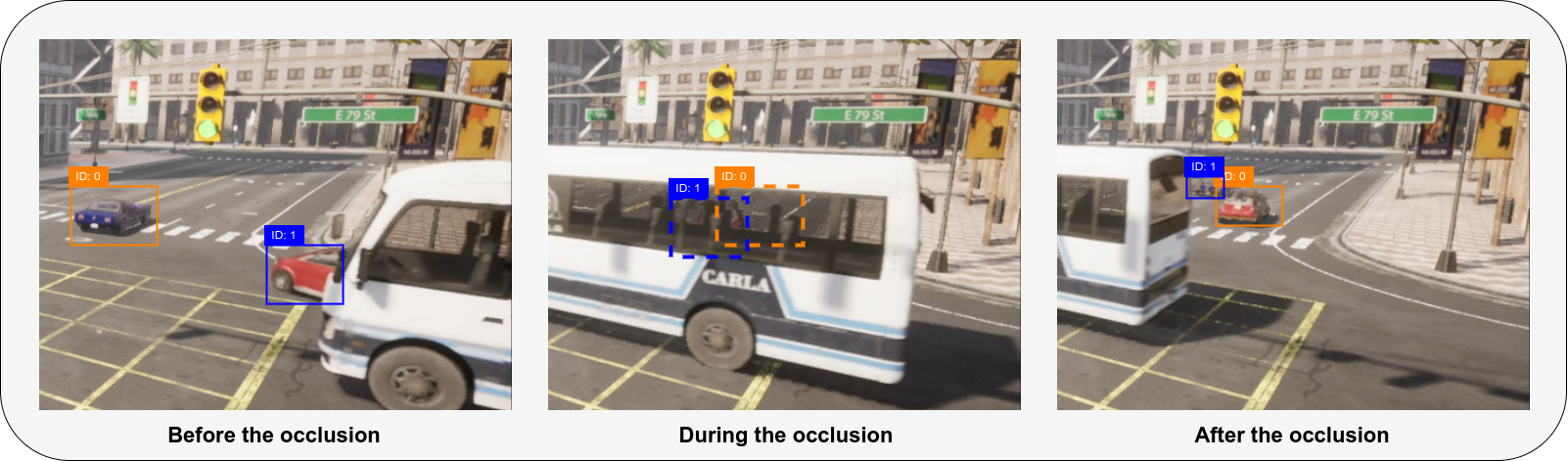}
  \caption{An example of id-switch caused by the use of just the IoU metric, which can be avoided with a neural re-identification network and the cosine similarity.}
  \label{fig:occl}
  \end{center}
\end{figure}

Consider, for example, the scenario represented in the three frames reported in Fig.~\ref{fig:occl}.

In this case, because of the progressive overlap, the position of the two objects becomes rapidly imprecise due to the absence of a measure to correct the Kalman filter prediction. During this transient, the bus can still be matched, while the the two cars under occlusion are updated by dead reckoning. When the occluded objects becomes visible again, they reappear in a position that is far from the last detection. Since BYTEtrack only uses the IoU match, the chance of missing the recovery of the track is high, especially for small objects. In fact, if the motion of the hidden objects results to be highly non-linear during the occlusion, its predicted position is likely to have a negligible IoU with the detected one or a high IoU with the wrong track, possibly causing an id-switch. In contrast, the solution implemented in CORT can recover the track in the second stage of the cascade matching, because it exploits the features extracted by the re-identification network. Then, using the cosine similarity, it is able to get a more reliable association score, so recognizing a previously occluded object, rather than creating a new track. The situation described above is very common in multi-object tracking and the results reported in Table~\ref{t:mot_score} confirm that the proposed solution improves the overall performance with respect to BYTEtrack by significantly reducing the number of id switches, with a small increase in the inference time.

\begin{table}[h!tbp]
\centering
\footnotesize
\begin{tabular}{ | c | c | c |  c | c |}
    \hline
    \textbf{Tracker} & \textbf{FPS}  & \textbf{IDS} & \textbf{MOTA} & \textbf{MOTP}\\ \hline
    Proposed solution & 23.8 &  3004 & 72.30 & 79.13 \\ \hline
    BYTEtrack  & 19.1 & 22989 & 62.79 & 80.33 \\ \hline
\end{tabular}
\vspace{2mm}
\caption{A comparison of the BYTEtrack and the proposed class-based solution using different evaluation metrics. }
\label{t:mot_score}
\end{table}

For the sake of fairness, the results reported in Table~\ref{t:mot_score} have been obtained by replacing the detection network of BYTEtrack with the one used in the proposed architecture, which is a YOLOv8 trained on the COCO dataset~\cite{coco}. In this way, both trackers worked with the same set of detections. The achieved results also show that the re-identification stage slightly increases the execution time of the proposed tracker with respect to BYTEtrack. To evaluate such a difference, however, one should consider that BYTEtrack is optimized for tracking multiple objects belonging to a single class, while the proposed model is optimized for multi-classes tracking. MOTA and MOTP were selected as evaluation metrics because they directly capture CORT’s goals of identity consistency and precise localization in multi-class tracking. MOTA consolidates tracking errors, like ID switches and occlusions, providing a straightforward measure of identity handling, while MOTP evaluates the alignment of tracked positions with the ground truth. HOTA, instead, is designed to capture long-term re-identification relations, which are not prioritized in the proposed solution.

\subsection{Evaluation on KITTI 2D}

The aim of this experiment is to show the results obtained by CORT on a standard multi-class multi-object tracking benchmark, like KITTI~\cite{KITTI}. The KITTI dataset includes seven different classes, but the benchmark returns the score evaluation on only two of them: car and person. This choice was made by the benchmark creators because these two classes were considered to have a sufficient number of occurrences to ensure a reliable evaluation. Table~\ref{t:kitti_score} reports the scores obtained for the two classes by the proposed solution (CORT) and DeepSORT.

\begin{table}[h!tbp]
\centering
\footnotesize

\textbf{Results for class Car} \\[5pt] 
\begin{tabular}{ | c | c | c |  c | c | c | }
    \hline
    \textbf{Tracker} & \textbf{IDS} & \textbf{MOTA} & \textbf{MOTP} & \textbf{HOTA} & \textbf{FPS}\\ \hline
     
    CORT (two stages) & 184 \textcolor{green}{\(\uparrow\)} & 85.66 \textcolor{green}{\(\uparrow\)} & 86.99 \textcolor{green}{\(\uparrow\)} & 72.64 \textcolor{green}{\(\uparrow\)} & 27.9 \\ \hline

    CORT (IoU only) & 256 \textcolor{red}{\(\downarrow\)} & 73.51 \textcolor{red}{\(\downarrow\)} & 82.51 \textcolor{green}{\(\uparrow\)} & 70.31 \textcolor{green}{\(\uparrow\)} & 30 \\ \hline
    
    DeepSORT & 237  & 78.14  & 76.66 & 59.98  & 19.7 \\ \hline
    
BYTEtrack & 242 & 76.11 & 78.37 & 71.63 & 24.8\\ \hline
    
\end{tabular}

\addvspace{12pt} 

\textbf{Results for class Person} \\[5pt]  
\centering
\footnotesize
\begin{tabular}{ | c | c | c |  c | c | c | }
    \hline
    \textbf{Tracker}  & \textbf{IDS} & \textbf{MOTA} & \textbf{MOTP} & \textbf{HOTA} & \textbf{FPS}\\ \hline
    CORT (two stages) & 263 \textcolor{green}{\(\uparrow\)} & 50.94  \textcolor{green}{\(\uparrow\)} & 66.45 \textcolor{green}{\(\uparrow\)} & 43.98 \textcolor{green}{\(\uparrow\)} & 27.2 \\ \hline

    CORT (IoU only) & 348 \textcolor{red}{\(\downarrow\)} & 43.56 \textcolor{red}{\(\downarrow\)} & 57.85 \textcolor{red}{\(\downarrow\)} & 38.96 \textcolor{green}{\(\uparrow\)} & 30 \\ \hline
    
    DeepSORT & 312 & 45.90 & 63.63 & 41.10 & 18.3 \\ \hline

BYTEtrack & 324 & 44.82 & 62.43 & 41.62 & 24.2 \\ \hline
    
\end{tabular}

\caption{Evaluation results on the Kitti 2D multi-object tracking benchmark.}
\label{t:kitti_score}
\end{table}

The proposed solution demonstrates significant improvements in both MOTA and MOTP metrics compared to DeepSORT, achieving higher tracking accuracy and localization precision.
For the Car class, the proposed solution achieves a MOTA of 85.66 and MOTP of 86.99, outperforming DeepSORT's respective scores of 78.14 and 76.66.
Similarly, for the Person class, the proposed method achieves a MOTA of 50.94 and MOTP of 66.45, confirming better results with respect to DeepSORT's scores of 45.90 and 63.63.

CORT also demonstrates a balanced performance in terms of HOTA, outperforming DeepSORT and showing strengths in maintaining tracking consistency for dynamic objects.
In particular, note that the HOTA improvement achieved by CORT over DeepSORT is higher for the Car class, confirming the effectiveness of the proposed solution in tracking objects with faster dynamics, whereas people are easier to re-associate after occlusions for both models.

We also carried out an ablation study to evaluate the effect of the second stage (in charge of re-identification) on the tracking speed and performance of CORT. The results reported in Table~\ref{t:kitti_score} remark the importance of the second stage for both reducing the number of ID switches (IDS) and recovering from temporary occlusions. 
Specifically, for the Car class, the IDS is reduced from 256 to 184, while MOTA improves from 73.51 to 85.66, and HOTA from 70.31 to 72.64. 
Similarly, for the Person class, the IDS is reduced from 348 to 263, with corresponding improvements in MOTA (from 43.56 to 50.94) and HOTA (from 38.96 to 43.98). 
These improvements come at the cost of a slight reduction in the frame rate, which decreases from 30 fps to 27.9 fps for the Car class and from 30 fps to 27.2 fps for the Person class. This feature provides an additional flexibility to the user for configuring the CORT pipeline depending on the requirements of the specific application.
A direct comparison with BYTEtrack highlighted CORT’s advantages in both accuracy and efficiency. For the Car class, CORT achieved higher HOTA (72.64 vs. 71.63) and MOTA (85.66 vs. 76.11), with fewer ID switches (184 vs. 242) and a slightly higher FPS (27.9 vs. 24.8). Similarly, for the Person class, CORT improved HOTA (43.98 vs. 41.62) and MOTA (50.94 vs. 44.82), reducing ID switches (263 vs. 324) while maintaining a higher FPS (27.2 vs. 24.2). These results confirm that the proposed multi-class association and cascade matching strategies enhance both tracking robustness and computational efficiency.

In summary, these results highlight the robustness of the proposed solution in achieving precise object localization and reducing ID switches in challenging multi-class scenarios, without penalizing real-time performance.

\subsection{Behavior under unstable detections}
\label{unst}

\begin{figure}[h!]
  \begin{center}
  \includegraphics[width=1.0\linewidth]{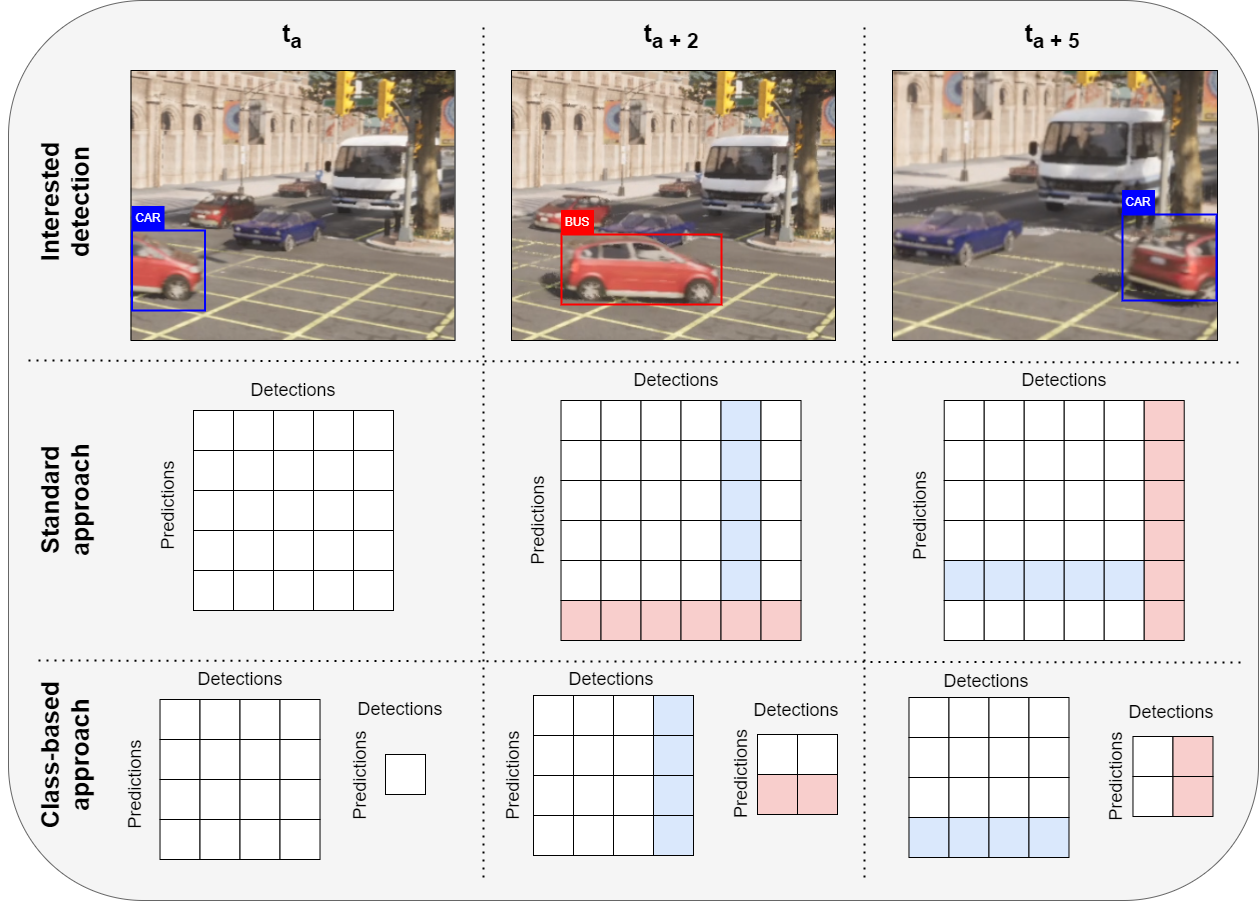}
  \caption{Evolution of the Hungarian matrices in the presence of unstable detections for the standard monolithic approach and the proposed multi-class solution.}
  \label{fig:unstable_dets1}
  \end{center}
\end{figure}

This section examines the behavior of the standard and the multi-class approach in two critical situations in which the detection network could provide sporadic erroneous outputs in a frame sequence.
To show the potential impact of such events on the performance of the tracker, consider a scenario similar to the one illustrated in Fig.~\ref{fig:unstable_dets1}, where the red car appears in the scene at time $t_a$, but it is misclassified as a bus at time $t_a + 2$. Also assume that all the other objects are correctly classified across the sequence. For the sake of clarity, only the detection responsible for the problem is shown in all the frames.  

To better understand the negative impact of such a misclassification, Fig.~\ref{fig:unstable_dets1} also reports the evolution of the Hungarian matrices for both the standard monolithic approach and the proposed class-based solution, highlighting for each frame how this specific detection affects the size of the matrices and, hence, the number of operations required to fill them.
In particular, the colored rows represent the elements required in the cost matrix to include the track that must be created to match the missing detection, while the colored columns represent dummy values that need to be created in the cost matrix to meet the squareness constraint imposed by the Hungarian match. The red and blue colors are associated with the two different classes reported in the experiment. In particular, the red color is associated with the Bus class while the blue color is associated with the Car class.

At time $t_{a}$, there are five predictions and five detections assumed to be correctly tracked.

Under this assumption, at time $t_{a+2}$, when the car is misclassified as a bus, one track is missing a corresponding detection and there is a new detection that requires the creation of a new tentative track. To correctly create the entry for the new track, cost matrix must have dimension $6 \times 6$ and, in such a context, the red row represents the space needed for creating the new track, while the blue column represents the space needed for creating a dummy detection for the car track at frame $t_{a}$, which now has no corresponding detection and is updated by using the Kalman filter prediction. The same rows and columns are reported also for CORT's approach, which is clearly more resource efficient. Assuming an \textit{in\_patience} value of 5 frames, then at time $t_{a+5}$, when the car is again correctly recognized, the spatial resolution of the matrices does not change, meaning that the proposed approach not only benefits from the parallel computation, but also takes advantage of a better resources usage, reducing the number of operations to be executed within each thread. The transient between [$t_{a+2}$; $t_{a+5}$] shows that the just mentioned improvements not only apply for the frame where the misclassification occurs, but also for a time interval that depends on the tracking hyperparameters discussed in Section~\ref{s:prop_pipe}. 

A simple solution to attenuate the two problems described in this section is to increase the acceptance threshold of the detection network. Raising the value of such a threshold ensures that only reliable detections are passed to the tracker to compute the costs for the assignment problem, and it is very unlikely to have unstable detections with a high confidence score. At the same time, raising the threshold increases the number of false negatives. However, this is not a major issue, because the Kalman filter can efficiently compensate for some missing detections once the track is promoted effective, as explained in Section~\ref{s:prop_pipe}. This is the reason why in Table~\ref{t:mot_score} the value for the precision (MOTP) is lower than that of the BYTEtrack, while the accuracy (MOTA) is higher because of the reduced number of id-switches.

\section{Conclusion and Future work}
\label{s:concl}

This work presented CORT, an efficient method to enhance the timing performance of modern real-time tracking algorithms. Such an enhancement is obtained by splitting the Hungarian matching algorithm into smaller association problems, carried out per class. Considering that the matching algorithm has an $O(n^3)$ time complexity, where $n$ is the number of elements in the cost matrix, and that each sub-matrix can be solved in parallel on different cores, the proposed approach significantly improves the performance of state-of-the-art tracking-by-detection systems. In real-world applications, such as autonomous vehicles and surveillance, high-speed tracking of CORT enables a rapid and reliable identification of multiple objects belonging to different classes. This capability supports real-time decision-making in complex environments, where stable execution times and reduced resource demands are crucial. As discussed in Section~\ref{s:carla_exp}, when the tracker operates in scenarios with a large number of objects, the proposed parallel solution demonstrates clear advantages in execution speed and scalability over monolithic approaches. This is also confirmed by the experimental results reported in Table~\ref{t:ex_time}, which show that the gap between the execution time of the two methods significantly increases in scenarios with a larger number of objects, underscoring the superior scalability of the proposed approach.

As a future work, we plan to further improve efficiecy by accelerating some tensor operations on FPGA, as those required by the Kalman filter and the Hungarian algorithm. Direct communication between modules on the FPGA will allow these tasks to be offloaded from the CPU, achieving more stable execution times and higher system throughput. Another optimization will be the dynamic tuning of the tracker's hyperparameters. For example, re-identification thresholds could be adjusted in real-time based on scene complexity or object density, enabling the tracker to automatically adapt to changing conditions without manual intervention.


\section*{Acknowledgments}
This work was supported by the Department of Excellence of Robotics and AI, Scuola Superiore Sant'Anna, Pisa, Italy.

\def\refname{References}
\bibliographystyle{IEEEtran}
\bibliography{bibliography}

\begin{IEEEbiography}[{\includegraphics[width=1in,height=1.25in,clip,keepaspectratio]{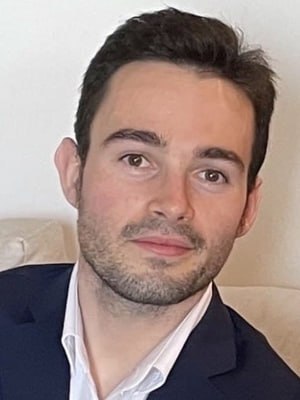}}]{Edoardo Cittadini}
received the master’s degree (cum laude) in embedded computing systems engineering jointly offered by the Scuola Superiore Sant’Anna of Pisa, Pisa, Italy, and the University of Pisa, Pisa, in 2021. He is currently pursuing the Ph.D. degree with the Real-Time Systems (ReTiS) Laboratory, Scuola Superiore Sant’Anna of Pisa. His research interests include cyber-physical systems, artificial intelligence on heterogeneous platforms, real-time object tracking, and hardware acceleration.
\end{IEEEbiography}

\begin{IEEEbiography}[{\includegraphics[width=1in,height=1.25in,clip,keepaspectratio]{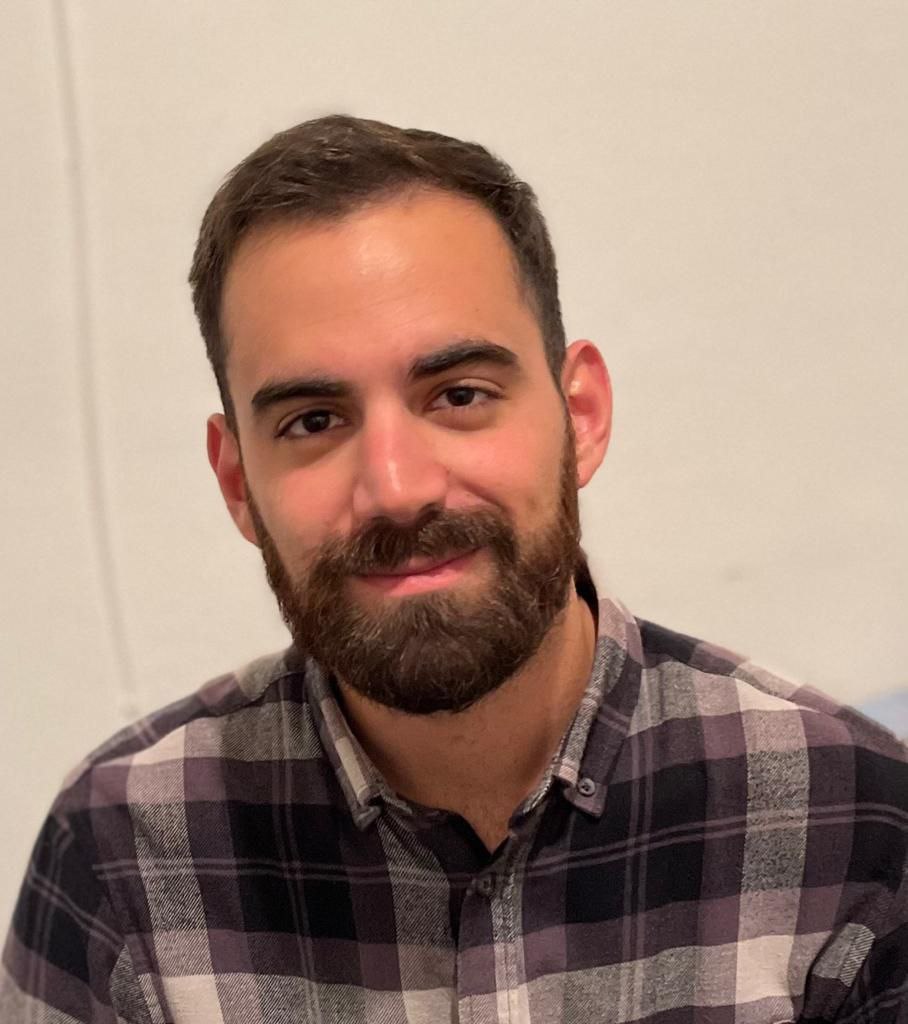}}]{Alessandro De Siena}
received his master's degree in Robotics and Automation Engineering at the University of Pisa in 2023. His thesis on multiple object tracking involved an in-depth examination of topics such as artificial intelligence, trajectory estimation algorithms, and autonomous driving vehicles. He implemented real-time tracking systems for UAV and developed object detection models for embedded devices. Currently, he has a position at Softsystem, an automation company located in Pisa.
\end{IEEEbiography}

\begin{IEEEbiography}[{\includegraphics[width=1in,height=1.25in,clip,keepaspectratio]{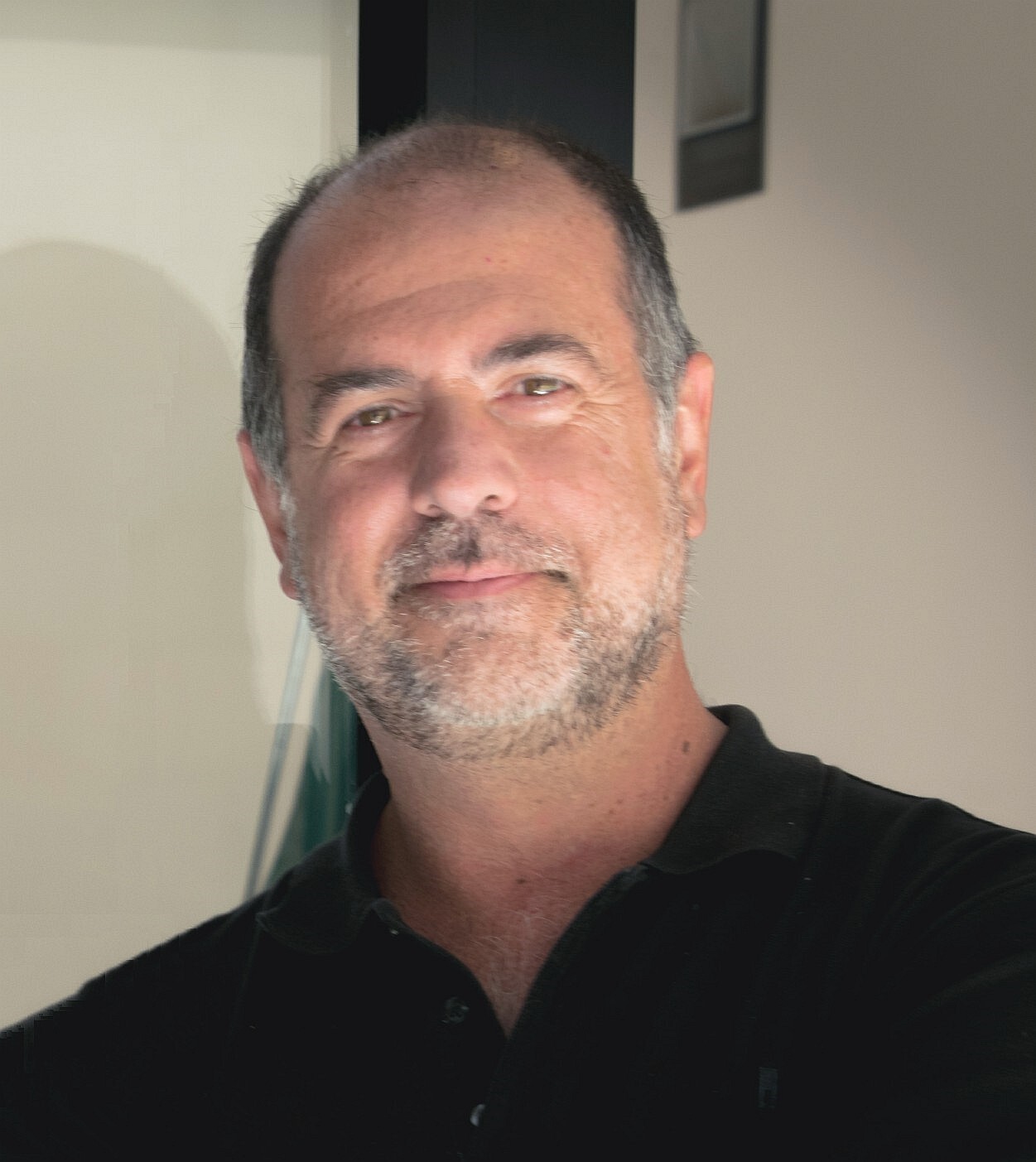}}]{Giorgio Buttazzo}
is full professor of computer engineering at the Scuola Superiore
Sant'Anna of Pisa. He graduated in Electronic Engineering at the University of Pisa,
received a M.S. degree in Computer Science at the University of Pennsylvania, and a
Ph.D. in Computer Engineering at the Scuola Superiore Sant'Anna of Pisa.
He has been Editor-in-Chief of Real-Time Systems, Associate Editor of IEEE Transactions
on Industrial Informatics and ACM Transactions on Cyber-Physical Systems. He is IEEE fellow
since 2012, wrote 7 books on real-time systems and more than 300 papers in the field of
real-time systems, robotics, and neural networks, receiving 15 best paper awards.
\end{IEEEbiography}

\end{document}